# Revolutionizing Underwater Exploration of Autonomous Underwater Vehicles (AUVs) and Seabed Image Processing Techniques

**Dr. Rajesh Sharma R**
Associate Professor,
Department of Computer Science and Engineering,
Alliance College of Engineering and Design,
Alliance University, Bangalore, India.
rajeshsharma.r@alliance.edu.in

**Dr. Akey Sungeetha**
Associate Professor,
Department of Computer Science and Engineering,
Alliance College of Engineering and Design,
Alliance University, Bangalore, India.
akey.sungheetha@alliance.edu.in

**Dr. Chinnaiyan R**
Professor,
Department of Computer Science and Engineering,
Alliance College of Engineering and Design,
Alliance University, Bangalore, India.
chinnaiayan.r@alliance.edu.in

*Abstract-* The oceans in the Earth's in one of the last border lines on the World, with only a fraction of their depths having been explored. Advancements in technology have led to the development of Autonomous Underwater Vehicles (AUVs) that can operate independently and perform complex tasks underwater. These vehicles have revolutionized underwater exploration, allowing us to study and understand our oceans like never before. In addition to AUVs, image processing techniques have also been developed that can help us to better understand the seabed and its features. In this comprehensive survey, we will explore the latest advancements in AUV technology and seabed image processing techniques. We'll discuss how these advancements are changing the way we explore and understand our oceans, and their potential impact on the future of marine science. Join us on this journey to discover the exciting world of underwater exploration and the technologies that are driving it forward.

*Keywords: AUV, ROV, sonar imaging, underwater exploration.*

## 1. INTRODUCTION

Introduction to Autonomous Underwater Vehicles (AUVs) In recent decades, the field of underwater exploration has witnessed a remarkable transformation with the advent of Autonomous Underwater Vehicles (AUVs) [1]. These cutting-edge technological marvels have revolutionized the way scientists, researchers, and explorers delve into the mysteries of the deep sea [2]. Autonomous Underwater Vehicles, commonly known as AUVs, are unmanned robotic vehicles designed to navigate and operate independently in the vast and challenging underwater environment. Unlike remotely operated vehicles (ROVs) that require human intervention and control, AUVs are self-guided and capable of executing pre-programmed missions with remarkable precision. These underwater drones are equipped with a plethora of advanced sensors, cameras, and scientific instruments, allowing them to collect a wealth of data and capture high-resolution images of the seabed with unparalleled accuracy. By venturing into the depths where humans cannot reach, AUVs provide a unique opportunity to explore uncharted territories, study marine ecosystems, map underwater geological features, and even search for sunken treasures. The development and deployment of AUVs have significantly enhanced our understanding of the marine environment, unlocking new possibilities for scientific research, environmental monitoring, resource exploration, and underwater archaeology. With their ability to operate autonomously for extended periods, [3] AUVs have proven to be invaluable assets in gathering crucial data and conducting surveys in remote and challenging underwater locations. Furthermore, the integration of advanced seabed image processing techniques has further amplified the capabilities of AUVs [4]. By employing sophisticated algorithms and computer vision technology, these techniques enable the analysis and interpretation of the vast amount of visual data captured by AUVs. From identifying marine species and habitats to detecting geological formations and underwater artifacts, seabed image processing techniques provide invaluable insights into the underwater world. In this comprehensive survey, we will delve into the depths of the mesmerizing field in the Autonomous Underwater Vehicles (AUVs) and explore the fascinating realm of seabed image processing techniques. Join us as we embark on an exhilarating journey to uncover the secrets hidden beneath the ocean's surface and witness the remarkable advancements that are revolutionizing underwater exploration.

## 2. THE EVOLUTION OF AUVS

Over the years, there has been a remarkable evolution in the field of underwater exploration, particularly in the development of Autonomous Underwater Vehicles (AUVs). These cutting-edge machines have transformed the way we explore and study the depths of our oceans [5]. Initially, underwater exploration relied heavily on remote-controlled vehicles, which required constant human intervention and control. While these vehicles could capture valuable data and images from the seabed, their limitations became apparent as the need for more efficient and independent operations emerged [6]. The transition from remote-controlled vehicles to autonomous AUVs marked a significant milestone in underwater exploration. These advanced machines are equipped with sophisticated sensors, navigation systems, and artificial intelligence algorithms that enable them to operate independently, without human intervention [7],[8]. The autonomy of AUVs allows them to perform complex tasks, navigate through challenging underwater terrains, and collect data with precision and accuracy. They can be pre-programmed to follow specific routes, conduct surveys, and execute scientific experiments, all while adapting to changing environmental conditions [9]. This evolution in AUV technology has revolutionized underwater exploration by expanding the boundaries of our understanding of the marine world. These remarkable vehicles have become invaluable tools for oceanographic research, environmental monitoring, underwater archaeology, and offshore industries. Furthermore, the advancements in AUV technology have also led to the development of innovative seabed image processing techniques. These techniques involve the analysis and interpretation of images captured by AUVs, enabling researchers to extract valuable information about the underwater environment, marine life, and geological formations [10]. The combination of autonomous AUVs and advanced image processing techniques has opened new possibilities for underwater exploration, providing scientists and researchers with unprecedented access to the hidden depths of our oceans. In the next sections of this comprehensive survey, we will delve deeper into the capabilities of AUVs and explore the various seabed image processing techniques that are revolutionizing underwater exploration. Stay tuned for an in-depth exploration of these remarkable technologies and their impact on our understanding of the underwater world.

## 3. THE SIGNIFICANCE OF AUVS IN UNDERWATER EXPLORATION

Autonomous Underwater Vehicles (AUVs) have revolutionized the field of underwater exploration, unlocked new possibilities, and pushed the boundaries of our understanding of the deep-sea environment [11]. These advanced robotic vehicles have become indispensable tools for scientists, researchers, and marine explorers, offering a wide range of capabilities and applications. One of the key significances of AUVs lies in their ability to operate autonomously, without the need for direct human control or intervention. This autonomy allows AUVs to perform complex missions in remote and challenging underwater environments, where human presence would be impractical or even impossible [12]. By removing the limitations of human endurance and the risks associated with deep-sea exploration, AUVs have opened a whole new world of possibilities for scientific discoveries and data collection. AUVs are equipped with a suite of sensors, cameras, and instruments that enable them to capture high-resolution images, collect data on water properties, map the seafloor, and even conduct experiments. These vehicles are capable of diving to great depths, maneuvering through intricate underwater terrains, and navigating with remarkable precision. This enables scientists to gather valuable data on marine life, underwater ecosystems, geological formations, and other phenomena that were previously inaccessible or difficult to study. Furthermore, AUVs can be programmed to follow predetermined paths or execute specific tasks, allowing for efficient and targeted exploration [13]. They can be deployed in large numbers, working collaboratively to cover vast areas and provide a comprehensive understanding of the underwater environment. This scalability and adaptability make AUVs valuable assets in various fields, including marine biology, oceanography, geology, archaeology, and environmental monitoring. In addition to their exploration capabilities, AUVs have also played a crucial role in advancing seabed image processing techniques. By capturing high-resolution images of the seafloor and underwater landscapes, AUVs have provided researchers with a wealth of visual data that can be analyzed and processed to extract valuable insights. These image processing techniques include bathymetry, photogrammetry, and side scan sonar imaging, among others, which allow for detailed mapping, habitat assessment, and identification of submerged objects or structures.

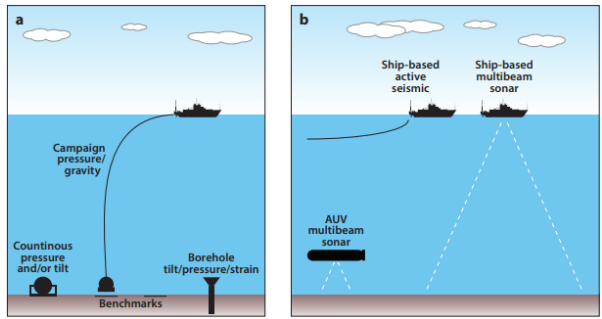

Figure 1. Sonar and Seismic Surveying

In the (a) Pressure sensors, it is deployed in in the ocean with campaigns or in permanent with connected wired mechanisms, to detect seafloor motions. Magnitude quantities are most important to vertical motions and sub-surface mass reorganizations. Ocean boreholes can be instrumented with tilt and strain meters, and pressure sensors. (b) Vessel-based sonar can achieve bathymetrical maps, that can be diverse for variation the measurements on the order of deep ocean. Sonar AUVs which permits higher-resolution bathymetry. Active-source seismic exploration from ships (sometimes aided by cabled seafloor geophone arrays) provides 4D time-lapse images of subsurface deformation [2]. In conclusion, AUVs have revolutionized underwater exploration by offering autonomy, versatility, and advanced sensing capabilities. These vehicles have expanded the scope of scientific study, enabling researchers to delve deeper into the mysteries of the ocean and uncover hidden treasures beneath the waves. With ongoing advancements in technology and the continuous refinement of AUV capabilities, we can expect even greater breakthroughs and discoveries in the realm of underwater exploration in the years to come.

## 4. EXPLORING THE VARIOUS TYPES OF AUVS AVAILABLE

When it comes to underwater exploration, Autonomous Underwater Vehicles (AUVs) have revolutionized the way scientists and researchers gather data from the depths of the sea [14],[15]. These unmanned vehicles are designed to operate independently, without the need for human intervention, making them ideal for conducting surveys in remote and hazardous underwater environments. There are various types of AUVs available, each with their own unique capabilities and functionalities. One such type is the glider AUV, which uses changes in buoyancy to propel itself through the water. Gliders are particularly useful for long-duration missions, as they can travel for weeks or even months, collecting data and transmitting it back to researchers in real-time.

Another type of AUV is the propeller-driven vehicle, which relies on the thrust generated by its propellers to navigate underwater [17]. These vehicles are typically faster and more maneuverable than gliders, making them ideal for tasks that require quick responses or precise movements. Additionally, there are hybrid AUVs that combine the features of gliders and propeller-driven vehicles, providing a balance between endurance and maneuverability [18]. These hybrid vehicles can switch between different modes of operation, allowing researchers to adapt to changing environmental conditions or specific mission requirements.

Furthermore, there are specialized AUVs designed for specific tasks, such as underwater mapping, seafloor imaging, or environmental monitoring. These vehicles are equipped with advanced sensors and imaging systems, enabling them to capture detailed data about the underwater environment, including seabed topography, water quality parameters, or the presence of marine organisms. Overall, the availability of various types of AUVs has opened new possibilities for underwater exploration and research. Researchers now have access to advanced technologies that can navigate the depths of the sea, collect valuable data, and provide insights into the mysteries of the underwater world.

## 5. UNVEILING THE CAPABILITIES OF AUVS IN SEABED IMAGE PROCESSING

Autonomous Underwater Vehicles (AUVs) have revolutionized the field of underwater exploration by offering unparalleled capabilities in seabed image processing. These advanced robotic systems are equipped with state-of-the-art sensors and imaging technologies that enable them to capture high-resolution images of the ocean floor with exceptional precision and clarity.
One of the key capabilities of AUVs lies in their ability to navigate autonomously through underwater environments, collecting vast amounts of data in the form of seabed images. These images provide valuable insights into the hidden world beneath the ocean's surface, allowing scientists, researchers, and marine biologists to study and analyze marine ecosystems, geological formations, and even archaeological sites with unprecedented detail [19],[20]. AUVs are equipped with advanced imaging systems, such as side-scan sonars, multibeam echosounders, and high-resolution cameras, which capture detailed images of the seabed. These images are then processed using sophisticated algorithms and techniques to enhance their quality and extract valuable information.
Seabed image processing techniques employed by AUVs involve various steps, including image enhancement, feature extraction, and classification. Image enhancement algorithms are employed to reduce noise, improve contrast, and enhance the overall quality of the images. This ensures that the resulting images provide a clear and accurate representation of the seabed. Feature extraction techniques play a crucial role in identifying and isolating specific objects or geological features of interest within the seabed images. These techniques utilize pattern recognition algorithms and machine learning approaches to detect and extract objects such as coral reefs, shipwrecks, and underwater geological formations.

Classification algorithms are then applied to categorize and classify the extracted features based on predefined criteria. This enables researchers to gain insights into the composition, distribution, and characteristics of different seabed elements, facilitating a deeper understanding of marine ecosystems and geological processes. The capabilities of AUVs in seabed image processing are truly remarkable, opening new avenues for scientific research and exploration. These autonomous underwater vehicles have the potential to uncover hidden treasures, unravel mysteries of the deep sea, and contribute to our understanding of Earth's underwater realms like never before. As technology continues to advance, we can expect even more groundbreaking developments in this field, further propelling the revolution of underwater exploration through AUVs and seabed image processing techniques.

## 6. ENHANCING IMAGE QUALITY AND RESOLUTION IN UNDERWATER ENVIRONMENTS

Enhancing image quality and resolution in underwater environments has been a significant challenge for researchers and engineers in the field of autonomous underwater vehicles (AUVs) and seabed image processing. The underwater environment presents unique obstacles such as poor visibility, limited lighting conditions, and scattering of light, which can degrade the quality of images captured by AUVs. To overcome these challenges, several innovative techniques have been developed to enhance image quality and resolution in underwater environments. One of the most common approaches is the use of advanced imaging systems equipped with high-definition cameras and specialized optics designed specifically for underwater conditions. These systems often incorporate features such as image stabilization, color correction, and noise reduction algorithms to improve the clarity and detail of underwater images.
Another approach is the use of artificial lighting systems, such as strobe lights or LED panels, to provide additional illumination in dark or murky underwater conditions. These lighting systems can significantly enhance the visibility and contrast of underwater images, allowing for better identification and analysis of underwater features. In addition to hardware improvements, sophisticated image processing algorithms have been developed to address the challenges of underwater image enhancement. These algorithms utilize techniques such as dehazing, contrast enhancement, and image restoration to reduce the effects of light scattering and improve the overall quality and resolution of underwater images. Machine learning and computer vision techniques are also being employed to automatically detect and correct image distortions caused by factors such as water currents or camera movement.

The continuous advancements in imaging technology and image processing techniques are revolutionizing underwater exploration by enabling researchers and scientists to gather high-quality, detailed imagery of the underwater world. These advancements not only facilitate the study of marine ecosystems and underwater archaeological sites but also have practical applications in fields such as marine resource management, underwater inspection, and offshore oil and gas exploration. As the demand for underwater exploration and research grows, further advancements in image quality and resolution enhancement techniques are expected. These advancements will pave the way for new discoveries and a deeper understanding of the mysteries that lie beneath the surface of our oceans.

## 7. OVERCOMING CHALLENGES IN UNDERWATER DATA COLLECTION AND PROCESSING

Underwater exploration has always presented unique challenges when it comes to data collection and processing. The harsh underwater environment, limited visibility, and complex seafloor terrain make it difficult to gather accurate and reliable data. However, with the advancements in technology, autonomous underwater vehicles (AUVs) have revolutionized underwater exploration by overcoming these challenges. One of the major challenges faced in underwater data collection is the limited accessibility for humans. Diving to great depths is not only risky but also time-consuming [21]. AUVs, on the other hand, can be deployed for longer durations and reach depths that humans cannot. Equipped with sensors and cameras, these vehicles can collect a vast amount of data, including high-resolution images of the seafloor.

Once the data is collected, processing and analyzing it becomes another hurdle. Underwater images often suffer from poor visibility, distortion, and noise, making it difficult to extract meaningful information [22]. To overcome this, sophisticated image processing techniques have been developed specifically for underwater applications. These techniques include image enhancement, noise reduction, and image stitching, among others. Image enhancement algorithms are used to improve the visibility of underwater images by reducing color distortion and enhancing contrast. Noise reduction techniques, such as adaptive filters, are employed to remove unwanted artifacts and improve the overall quality of the images [23]. Additionally, image stitching algorithms are applied to seamlessly merge multiple images together, creating a more comprehensive and accurate representation of the seafloor.

Another challenge in underwater data processing is the vast amount of data collected by AUVs. Processing this large volume of data manually is not practical, which is why automated algorithms and machine learning techniques are utilized. These algorithms can efficiently analyze the data, identify objects of interest, and extract valuable insights. By overcoming these challenges in underwater data collection and processing, AUVs have paved the way for significant advancements in underwater exploration. Researchers and scientists can now gather precise and detailed information about the seafloor, marine life, and underwater ecosystems. This comprehensive understanding not only contributes to scientific research but also aids in various industries such as marine biology, geology, and offshore energy exploration. In conclusion, the development of AUVs and advanced image processing techniques has revolutionized underwater exploration. These technologies have enabled us to overcome the challenges of data collection and processing in the underwater environment, providing us with valuable insights into the unexplored depths of our oceans.

## 8. CUTTING-EDGE TECHNOLOGIES REVOLUTIONIZING AUVS AND SEABED IMAGE PROCESSING

In recent years, the field of underwater exploration has witnessed a remarkable transformation, thanks to cutting-edge technologies that have revolutionized Autonomous Underwater Vehicles (AUVs) and seabed image processing techniques. These advancements have not only enhanced the efficiency and accuracy of underwater surveys but have also opened new frontiers for scientific research, environmental monitoring, and commercial applications. One of the groundbreaking technologies that have propelled AUVs to new heights is artificial intelligence (AI). AI-powered AUVs are capable of autonomously navigating through complex underwater terrains, collecting vast amounts of data, and making real-time decisions based on the information gathered. These intelligent machines can adapt to changing environmental conditions, optimize their routes, and avoid obstacles, all while maximizing their data acquisition capabilities. This level of autonomy has revolutionized underwater exploration by significantly reducing human intervention and increasing operational efficiency [24].

Seabed image processing techniques have also witnessed remarkable advancements, enabling researchers to extract valuable information from underwater imagery with unprecedented precision. Image recognition algorithms, based on deep learning and computer vision, can now identify, and classify various underwater objects and organisms with remarkable accuracy. This has revolutionized marine biology, enabling scientists to study marine ecosystems in greater detail and understand the intricacies of underwater life. Furthermore, advancements in underwater communication technologies have played a pivotal role in enhancing the capabilities of AUVs and seabed image processing. High-speed acoustic modems and underwater wireless networks have enabled real-time data transmission, enabling researchers to monitor and control AUVs remotely. This has not only improved operational efficiency but has also enabled collaborative exploration, where multiple AUVs can work together, sharing information and coordinating their activities to achieve common objectives.

The integration of these cutting-edge technologies has paved the way for unprecedented discoveries in underwater exploration. From mapping unexplored underwater terrains to studying marine biodiversity and monitoring underwater infrastructure, AUVs equipped with advanced seabed image processing techniques are transforming our understanding of the world beneath the waves. As researchers continue to push the boundaries of technology, the future of underwater exploration looks promising. With ongoing advancements in AI, underwater imaging, and communication technologies, AUVs will continue to revolutionize the way we explore and understand the mysteries of the deep sea [25]. The possibilities are endless, and we are only scratching the surface of what these innovative technologies can achieve in the realm of underwater exploration.

## 9. REAL-WORLD APPLICATIONS OF AUVS IN SCIENTIFIC RESEARCH AND INDUSTRY

Autonomous Underwater Vehicles (AUVs) have revolutionized underwater exploration, enabling scientists and industries to delve deeper into the mysteries of the ocean. These remarkable machines have a wide range of real-world applications that have greatly contributed to scientific research and various industries. One of the significant applications of AUVs is in marine biology and ecology. Scientists can deploy AUVs to study marine life, their behaviors, and habitats in their natural environment. These vehicles can capture high-resolution images and videos of underwater ecosystems, providing valuable insights into biodiversity, species interactions, and the impact of human activities on marine environments. AUVs can also be equipped with sensors to measure water quality parameters, helping researchers monitor and understand the health of marine ecosystems.

AUVs are also extensively used in the field of oceanography. These vehicles can collect data on temperature, salinity, currents, and other essential oceanographic parameters, providing valuable information for climate studies, ocean circulation models, and weather forecasting. By autonomously navigating through the depths of the ocean, AUVs can cover large areas and collect data at various depths, enabling scientists to gather a comprehensive understanding of the ocean's dynamics and processes. Moreover, AUVs have found practical applications in underwater archaeology and exploration. With their ability to navigate autonomously and capture high-resolution images, these vehicles have been instrumental in discovering and documenting submerged archaeological sites and historical artifacts. AUVs equipped with advanced imaging and mapping technologies can create detailed 3D models of underwater structures, providing archaeologists with valuable insights into our past. In the oil and gas industry, AUVs have become indispensable tools for pipeline inspection and maintenance. These vehicles can inspect underwater infrastructure, detect leaks or damages, and provide real-time data to operators, ensuring the integrity and safety of underwater pipelines. AUVs equipped with advanced imaging techniques can also assist in underwater exploration for oil and gas reserves, reducing the need for extensive human intervention and minimizing risks associated with traditional methods.

The applications of AUVs are not limited to these fields alone. They have also been used in underwater mine detection, environmental monitoring, search and rescue operations, and even deep-sea exploration. These versatile vehicles have opened endless possibilities for exploring the underwater world, pushing the boundaries of human knowledge, and understanding. As technological advancements continue to enhance the capabilities of AUVs and improve seabed image processing techniques, the potential for further groundbreaking discoveries and advancements in scientific research and industry is immense. The future of underwater exploration looks promising, thanks to the remarkable capabilities of AUVs and their invaluable contributions to various sectors.

## 10. THE FUTURE OF AUVS: ADVANCEMENTS AND POTENTIAL ADVANCEMENTS IN UNDERWATER EXPLORATION

The future of Autonomous Underwater Vehicles (AUVs) holds immense potential in revolutionizing underwater exploration. As technology continues to advance at an exponential rate, exciting developments are on the horizon, pushing the boundaries of what is possible in the realm of underwater research and discovery. One significant area of advancement lies in the expansion of AUV capabilities. These vehicles are becoming more sophisticated, equipped with advanced sensors, improved maneuverability, and enhanced communication systems. This allows them to navigate complex underwater environments with greater precision and efficiency, collecting a wealth of data and imagery in the process.

Furthermore, advancements in power sources and energy management systems are enabling AUVs to operate for extended periods, venturing into deeper waters and covering larger areas. This opens up new possibilities for exploring previously uncharted territories, uncovering hidden ecosystems, and studying elusive marine species that dwell in the deep sea. In addition to technological improvements, collaborations between researchers, scientists, and engineers are fostering innovation in underwater exploration. These collaborations are breaking down barriers and encouraging interdisciplinary approaches to problem-solving, resulting in groundbreaking advancements in AUV technology. One potential advancement that has garnered attention is the integration of artificial intelligence (AI) and machine learning algorithms into AUV systems. By incorporating these technologies, AUVs can adapt and learn from their surroundings, making intelligent decisions in real-time. This has the potential to significantly enhance data processing capabilities, improve object recognition, and aid in the identification of underwater features and phenomena.

Another area of focus is the development of advanced imaging and seabed processing techniques. By utilizing high-resolution cameras, multi-beam sonar systems, and sophisticated image processing algorithms, AUVs can capture detailed images of the seafloor, analyze geological formations, and identify underwater habitats and archaeological sites. This wealth of information can contribute to our understanding of marine ecosystems, geological processes, and historical maritime artifacts. As we look to the future, the potential for AUVs in underwater exploration is limitless. From mapping unexplored regions of the ocean to studying marine life behavior and monitoring environmental changes, these autonomous vehicles are poised to transform our understanding of the underwater world. Through continued advancements in technology, collaboration, and innovation, AUVs will play a pivotal role in unraveling the mysteries that lie beneath the surface of our vast oceans.